\documentclass{article}
\usepackage{arxiv}
\usepackage[utf8]{inputenc}
\usepackage[T1]{fontenc}
\usepackage{hyperref}
\usepackage{url}
\usepackage{graphicx}
\usepackage{natbib}
\usepackage{doi}

\title{MFL Data Preprocessing and CNN-based Oil Pipeline Defects Detection}
\author{ \href{https://orcid.org/0000-0002-5468-4079}{\includegraphics[scale=0.06]{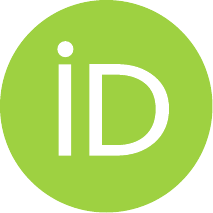}\hspace{1mm}Iurii Katser} \\
  Skolkovo Institute of Science and Technology\\
  Moscow, Russian Federation, 143026 \\
  \texttt{Iurii.Katser@skoltech.ru} \\
	\And
	\href{https://orcid.org/0000-0002-0770-9798}{\includegraphics[scale=0.06]{orcid.pdf}\hspace{1mm}Vyacheslav Kozitsin} \\
  Skolkovo Institute of Science and Technology\\
  Moscow, Russian Federation, 143026 \\
  \texttt{Vyacheslav.Kozitsin@skoltech.ru} \\
    \And
	\href{https://orcid.org/0009-0003-5436-6673}{\includegraphics[scale=0.06]{orcid.pdf}\hspace{1mm}Igor Mozolin} \\
	waico.tech\\
	Yerevan, Armenia, 0010 \\
	\texttt{mozolinia@gmail.com} \\
}

\hypersetup{
pdftitle={MFL Data Preprocessing and CNN-based Oil Pipeline Defects Detection},
pdfsubject={Deep learning, Computer vision, Convolutional neural networks, Anomaly detection, Fault detection, Oil pipelines, Magnetic Flux Leakage data, Defects, Technical diagnostics},
pdfauthor={Iurii Katser, Vyacheslav Kozitsin, Igor Mozolin},
pdfkeywords={Deep learning, Computer vision, Convolutional neural networks, Anomaly detection, Fault detection, Oil pipelines, Magnetic Flux Leakage data, Defects, Technical diagnostics},
}

\begin{document}

\maketitle

\begin{abstract}
Recently, the application of computer vision for anomaly detection has been under attention in several industrial fields. An important example is oil pipeline defect detection. Failure of one oil pipeline can interrupt the operation of the entire transportation system or cause a far-reaching failure. The automated defect detection could significantly decrease the inspection time and the related costs. However, there is a gap in the related literature when it comes to dealing with this task. The existing studies do not sufficiently cover the research of the Magnetic Flux Leakage data and the preprocessing techniques that allow overcoming the limitations set by the available data. This work focuses on alleviating these issues. Moreover, in doing so, we exploited the recent convolutional neural network structures and proposed robust approaches, aiming to acquire high performance considering the related metrics. The proposed approaches and their applicability were verified using real-world data.
\end{abstract}

\keywords{Deep learning \and Computer vision \and Convolutional neural networks \and Anomaly detection \and Fault detection \and Oil pipelines \and Magnetic Flux Leakage data \and Defect \and Technical diagnostics.}

\section{Introduction}
Anomaly detection problems have a great importance in industrial applications because anomalies usually represent faults, failures or the emergence of such \cite{Chandola2009}.
To detect them automatically, advanced analytic algorithms, including machine learning- and deep learning-based, can be applied. In this work, we investigated if deep neural network would perform well enough to provide hindsight to oil pipeline diagnostics. An oil pipeline system spans over thousands of kilometers, which makes manual inspection very costly and sometimes impossible. The damage of pipelines that transport oil and gas products leads to severe environmental problems. Eliminating leakages and their consequences is expensive.

\begin{figure}[ht]
	\center{\includegraphics[width=\textwidth]{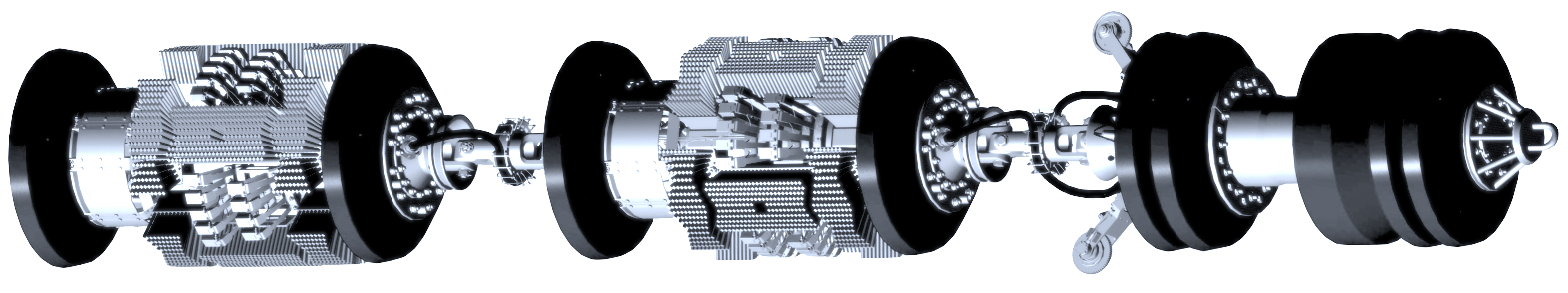}}
	\caption{In-line-inspection tool.}
	\label{ris:ili}
\end{figure}

To avoid accidents, it is recommended to improve the efficiency of diagnostics and increase the frequency of in-line-inspection (ILI) tools deployment (Fig.~\ref{ris:ili}).
ILI tools, also referred to as pipeline inspection gauges, use the Hall effect for measuring localized Magnetic Flux Leakage (MFL) intensity along the pipe wall. While moving along the pipe, the gauge inspects the wall and detects the magnetic field leaks. The MFL technique is the most common approach for nondestructive testing of oil and gas pipelines nowadays \cite{loskutov2006magnetic}.

The data collected during the inspection can be further analyzed and used to solve the main diagnostics problems \cite{katser2022machine}: detection of damages and defects, their localization, diagnosis or defects classification. 
Such analysis results are useful for assets management and repair prioritizing. This data analysis step is partly automated, but still there is a lot of manual work done here. That is why it is quite expensive and time consuming. Data analysis and machine learning techniques are very useful in the tasks of making processes more efficient time- and money-wise. Thus, an improved diagnostic process allows running the whole ILI procedure more often and gain more knowledge about the pipeline health, resulting in better safety and fewer financial losses due to leakages.

The objectives of this research are to appraise the proficiency of data engineering and computer vision (CV) techniques in oil pipeline diagnostics.

\section{Literature Review and Problem Statement}
\label{sec:litrev}

The MLF technique is the most common approach for nondestructive testing of oil and gas pipelines. The data obtained during the pipeline inspection is primarily analyzed by expert- and heuristics-based methods and since recently by regular machine learning (ML) methods. A comparison of performance among different ML methods for the defect identification problem is presented in \cite{Khodayari-Rostamabad2009}.
The main challenge for the ML approach is creating informative and important features that can be used as an input for ML methods. Usually, these diagnostic features are generated using expert knowledge and manually-created heuristics. So, on the one hand, the ML methods extend expert-based approaches and improve their quality. On the other hand, using manually generated features imposes the limitation on the quality of solving the ML-based defect detection problem that fails to fully automate the diagnostic process. A variety of most successful features is presented and analyzed in detail in \citet{Slesarev2017}.

To overcome the limitations of the expert-based and ML-based approaches, one can resort to Deep Learning (DL)  techniques that showed significant progress and achieved incredible results in numerous applications just in the past few years. The image classification problem is one of the most successful applications of DL and Convolutional Neural Networks (CNNs) in particular. CNNs can also be used to automate the process of feature generation in MFL data analysis. As an advantage, they can solve defect detection, weld strength detection, classification and segmentation problems at the same time. In literature there are examples of applying CNNs for defect detection \cite{Feng2017}, welds defect detection \cite{2020a}, weld and defects classification \cite{Yang2020}, and defect size estimation \cite{Lu2019}.
For all the mentioned applications, CNNs outperformed traditional approaches.

Nevertheless, still, there are few works dedicated to MFL data analysis using DL, and the existing DL approaches do not always achieve the required quality for full automation of the diagnostic process using such techniques. A number of particular problems that can be solved using the novel approach are not covered yet. For instance, we could not find any works on applying CNNs to the defect segmentation task, despite the importance of this problem according to \cite{Feng2017}. This can be an extension of the current research. This work seeks to address three different problems:
\begin{enumerate}
	\item Defect detection with the DL techniques,
	\item Welds strength detection with the DL techniques,
	\item MFL data preprocessing.
\end{enumerate}

To solve the first two problems, it is proposed to apply CNNs of different architecture and compare their results with the existing state-of-the-art approaches. It was decided to formulate such defect detection problem as an image classification problem in terms of ML because the applied DL techniques are intended to solve the problem formulated so. To solve the first problem, we state the binary  classification problem (healthy pipe or defected pipe). To solve the second problem, we state the multiclass classification problem (healthy pipe, defected pipe or defected weld), covering first two problems simultaneously.

Also, this research addresses different preprocessing techniques for dealing with typical issues in comparing the MFL data results of various preprocessing approaches, used with various CNNs. This work seeks to constructing such a preprocessing approach that improves the results of the defect detection problem best of all.

\section{Dataset Description}
\begin{figure}[ht]
	\center{\includegraphics[width=\textwidth]{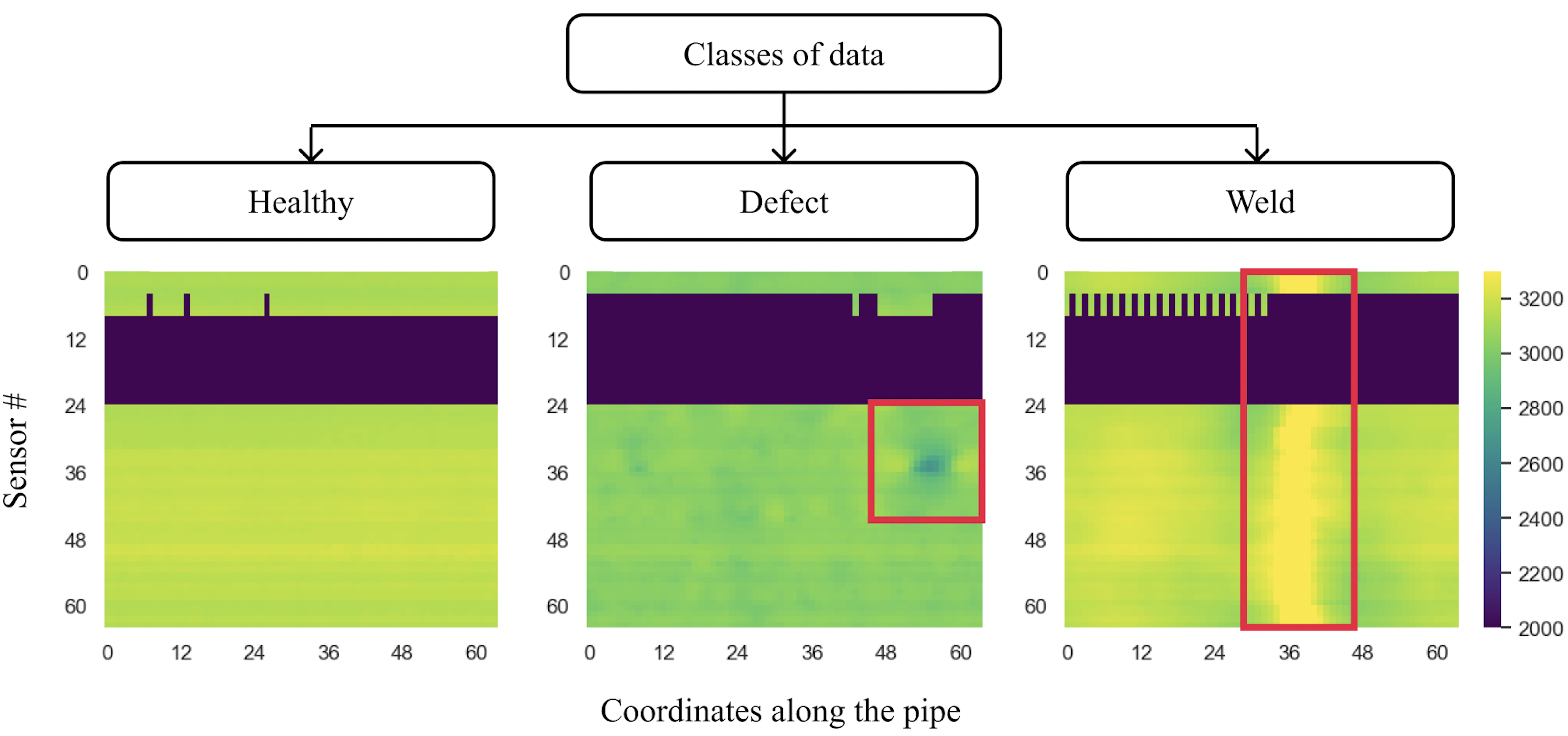}}
	\caption{Image classes distinguished in this work.}
	\label{ris:classes}
\end{figure}

There are three main classes of data that are attended to by diagnostic personnel.
They are presented in Fig.~\ref{ris:classes}.
Some other classes of data (concerning pipe tees and bends) are out of the scope of this work as well as different classes of defects and different classes of welds (healthy and defected).

Although MFL data looks quite similar for different pipes and ILI tools, it can differ significantly. The data mainly depends on pipe size, wall width, sensor geometry, and other geometric characteristics. Moreover, ILI tools differ a lot for different pipe sizes. Therefore, the repeatability of the results for different datasets should be investigated additionally. Further on, we provide dataset characteristics, which are also presented in Table~\ref{tab:dataset}.
The data was collected from a 219 mm in diameter pipe. The MFL dataset provides information about a single inspection tool run. The dataset has 64 features collected from 64 sensors installed at a constant step (10.75 mm) around the perimeter of the ILI tool. The data is collected as an array of 1x64 shape with a constant step (3.37 mm) along with the ILI tool movement inside the pipe. The dataset has 4,470,704 samples (steps along the pipe) that represent a 15,162.85 m part of pipeline. The sample values vary from 0 to 4,095 units. It has 745 defects of different types and 1,462 welds, 34 of which were found to be defected.
Figure~\ref{ris:classes} shows examples of healthy data, data with a weld, and with a defect. A technical report, attached to the dataset, contains information about the location of welds and defects, defect types, sizes, and other related characteristics. The report is prepared manually by the domain expert, so it contains some inaccuracies and needs additional preprocessing, as well as the data itself.
\begin{table}[!htb]
	\caption{Dataset characteristics}
	\begin{center}
		\small
		\begin{tabular}{ll}
			\hline
			Parameter & Value \\
			\hline
			Pipeline diameter, mm &  219 \\
			Pipeline length, m &  15162.85 \\
			Number of samples &  4470704 \\
			Number of features & 64 \\
			Min value & 0 \\
			Max value & 4095 \\
			Number of defects & 745 \\
			Number of welds (with defects) & 1462 (34) \\
			\hline
		\end{tabular}
		\label{tab:dataset}
	\end{center}
\end{table}

\section{Preprocessing Procedures}
Raw data has several issues that make it unusable to solve CV problems without proper preprocessing. The issues are:
\begin{enumerate}
	\item Sensor malfunctions (zeroed values cause bold horizontal line in Fig.~\ref{ris:classes}),
	\item Displaced origins between data and report coordinates,
	\item Inaccurate annotations, e.g., missed defects, wrong defect location, etc.,
	\item No annotated data for the segmentation task.
\end{enumerate}
The preprocessing stages and procedures that resolve these and other issues are given in this section.

\begin{figure}[ht]
	\center{\includegraphics[width=\textwidth]{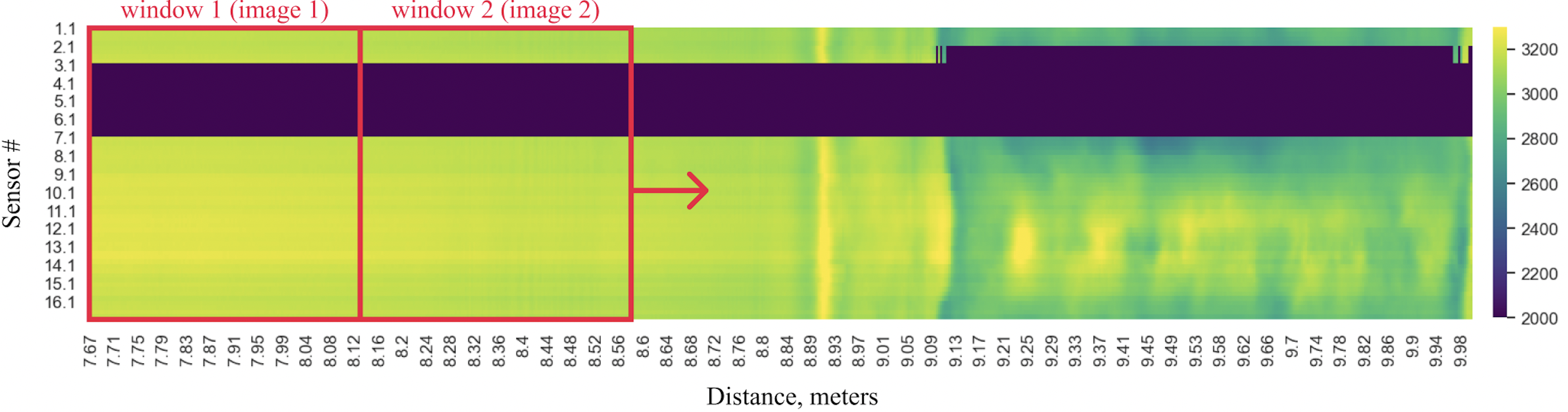}}
	\caption{Non-overlapping sliding window scheme for data preprocessing.}
	\label{ris:sw}
\end{figure}

\paragraph{Initial dataset transforming into separate images}
The initial dataset represents a long table indexed over the coordinate along the pipe. To state and solve the image classification problem, we should first decompose this long table into smaller squared 64x64 subtables. It can be also interpreted as a sliding window that runs over the coordinate (index) of the initial dataset and clips the dataset into the non-overlapping subsets (figure~\ref{ris:sw}). Each subset can be shown as an image of the pipe part. as a result, we had a dataset of 11,690 images of the healthy class, 711 images of the defected class, and 1,412 images with welds. The characteristics of the pipeline defect dataset are described in Table \ref{tab:alg1}. These classes were assigned to the images according to the markup from the technical report, where coordinates of the welds and defects are noted. Thus, the image covering the range of coordinates with the defect is interpreted as an image with the defected class. From now on, we refer to this transformed dataset of 13,813 64x64 images.

\paragraph{Sensors malfunctions problem}
To deal with sensor malfunctions, we propose to fill the gaps (zeroed values) with values calculated by different methods. Additionally, we will consider the values below 2,000 abnormal in this domain according to the experts and replace them with zeroes during the preprocessing.
\begin{enumerate}
	\item Abnormal values are equal to 0. Then Min-Max scaling to $[0.5:1]$ range.
	\item Abnormal values are equal to the mean of normal values from one picture. Then Min-Max scaling.
	\item Abnormal values are equal to the mean of normal values over the column. Then Min-Max scaling.
	\item Abnormal values are equal to the mean of neighboring sensors over the column. Then Min-Max scaling.
	\item Abnormal values are equal to the interpolation results over the column. Then Min-Max scaling.
\end{enumerate}

\begin{figure}[ht]
	\center{\includegraphics[width=\textwidth]{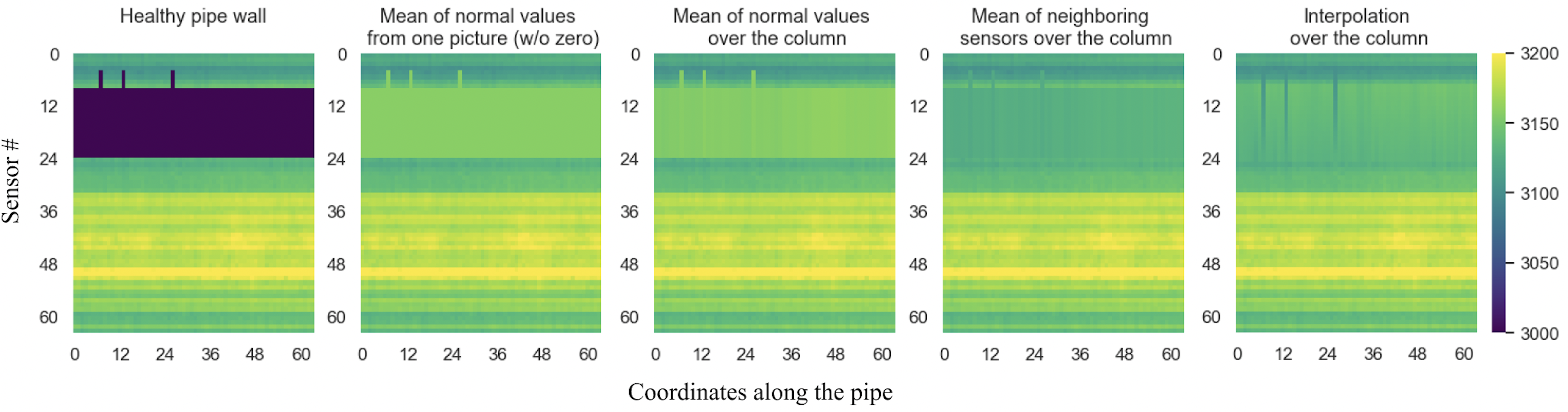}}
	\caption{Comparison of methods for missing values filling.}
	\label{ris:filling_example}
\end{figure}

The results of all the applied methods are presented in Fig.~\ref{ris:filling_example}. The Min-Max scaling can be applied using the whole dataset or just one image. Both approaches can be compared when the experiment is conducted.

Since the ILI tool location data did not match the defect location data from the report, it was necessary to merge the data. The key factor here turned out to be that the signal values from the magnetic flux sensors grew at the weld site. Hence the solution was to find the locations of the maxima of sensors data values and then to combine it with the weld coordinates.

\paragraph{Inaccurate annotations problem}
This problem is a common one for nondestructive testing of oil and gas pipelines \cite{Khodayari-Rostamabad2009}, as well as for the manual labelling.
There appears to be a lot of missing defects that affect the quality of the problem. Besides, there are wrong defect types and locations. To eliminate the wrong location issue, we additionally searched extremums around the provided location and chose the defects or welds, taking into account new coordinates.

\begin{figure}[ht]
	\center{\includegraphics[width=\textwidth]{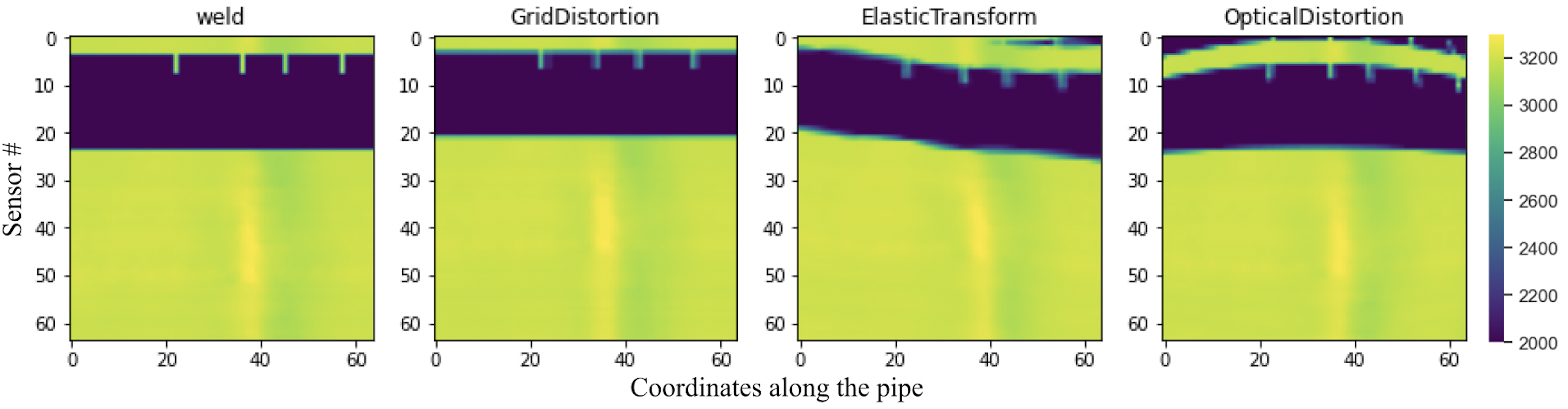}}
	\caption{Examples of augmented weld images.}
	\label{ris:aug_example}
\end{figure}

\paragraph{Augmentation}
Although we had a lot of data, we had small amount of defects and welds in comparison with healthy pipe wall instances. The augmentation procedure was used to balance the classes of images and improve the model quality by increasing the number of images in small classes (defects, welds). The Albumentations library \cite{buslaev2020albumentations} was selected as an augmentation tool.
All the applied augmentations both for welds and defects are presented in Table~\ref{tab:aug}. Based on domain knowledge, not all selected augmentations were applied to images with welds.
The applied augmentation details are presented in \cite{buslaev2020albumentations} and references therein. The characteristics of the augmented dataset, used for the research, are described in Table \ref{tab:alg1}. Examples of augmentations are shown in Fig.~\ref{ris:aug_example}.

\begin{table}[!htb]
	\caption{Applied augmentations}
	\begin{center}
		\small
		\begin{tabular}{lcc}
			\hline
			Augmentation Type & Welds & Defects \\
			\hline
			Rotate90 &  -- & + \\
			Rotate180 & + & + \\
			Rotate270 &  -- & + \\
			VerticalFlip & + & + \\
			HorizontalFlip & + & + \\
			ElasticTransform & + & + \\
			GridDistortion & + & + \\
			OpticalDistortion & + & + \\
			Transpose & -- & + \\
			RandomRotate90 &  -- & + \\
			\hline
		\end{tabular}
		\label{tab:aug}
	\end{center}
\end{table}

\begin{table}[!htb]
	\caption{\label{tab:alg1}Dataset size before and after augmentation}
	\begin{center}
		\small
		\begin{tabular}{l ccc}
			\hline
			Data & Healthy & Defect & Weld \\
			\hline
			\multicolumn{4}{c}{Before augmentation}  \\
			\hline
			Train  & 11106 & 569 & 1130 \\
			Validation & 584 & 142 & 282 \\
			\hline
			\multicolumn{4}{c}{After augmentation}  \\
			\hline
			Train  & 11106 & 8535 & 11300 \\
			Validation & 584 & 142 & 282 \\
			\hline
		\end{tabular}
	\end{center}
\end{table}

\section{Defects Detection Methods}
\label{sec:methods}
The Pipeline defect detection is composed of two problems. First, the defect should be detected, and second, it should be evaluated using the segmentation results. We propose here a novel CNN architecture for solving the first problem. Additionally, we present the existing architectures that achieve the best results in the MFL and X-ray defect detection problems. 

\subsection{CNN Preliminaries}
A CNN is a special type of a neural network that has proven  effective in computer vision applications. State-of-the-art results can be achieved in the segmentation and classification tasks \cite{a10}. Compared to the computer vision algorithms that do not take advantage of CNNs, much less pre-processing is required. More importantly, such networks are able to learn characteristics from data, which otherwise would have to be individually accounted for \cite{a11}.

Even though CNNs have been proposed in different architectures - to increase their efficiency for specific tasks and/or datasets, only three types of layers are used without exception, each with a specific propose. They are convolutional, pooling, and fully connected (linear) layers. The convolutional layers aim to extract feature maps of the input images by applying filters over different region of images. For instance, with $k$ filters, each filter having weight and bias of $w_i$ and $b_i$, respectively, the convolution of an image patch, $x_n$, can be written as follows:

\begin{equation}
f_{i,n}=\sigma(W_ix_n+b_i),
\end{equation}

where $\sigma$ is the activation function. Besides the rectified linear units (ReLU), sigmoid or softmax activation functions, a multitude of different options exist, all having their individual advantages. These are applied on the layers’ output neurons (e.g. after a convolutional layer). After a number of convolutional layers, the pooling layers are commonly applied in prominent network architectures to reduce the size of particular dimensions. Max-pooling and average-pooling are two examples. The pooling layers, alongside the reducing dimension sizes, perform denoising when utilized on images. The fully connected layers are generally the last layers of CNNs, possessing a similar structure compared to the traditional neural networks \cite{a12}.

\subsection{Existing CNNs}
We implemented the CNN from \cite{Feng2017} with only one difference: we used squared pictures (64x64 pixels) as an input, so we omitted the Normalization layer. The interested reader can find all details and overall architecture parameters in \cite{Feng2017}.
From now on, this CNN is marked as CNN-2 by the number of Convolutional layers. We also implemented the CNN from \cite{2020a}, which showed better results than the pretrained one, and fine-tuned OverFeatNet, VGGNet, and GoogleNet networks. Since our input size was smaller than in the original paper, we used smaller kernel size (3x3 instead of 7x7). All the details and CNN parameters are presented in \cite{2020a}.
From now on, this CNN is marked as RayNet.

\subsection{Proposed CNN-5 model}
\begin{figure}[ht]
	\center{\includegraphics[width=\textwidth]{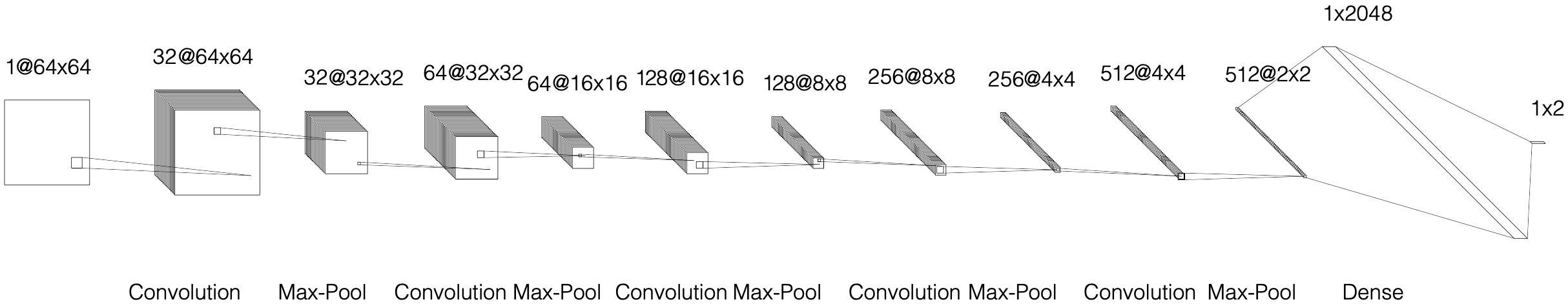}}
	\caption{Proposed CNN architecture.}
	\label{ris:CNN_our}
\end{figure}

The proposed model in Fig.~\ref{ris:CNN_our} consists of five convolutional layers overall. Each convolutional layer is followed by BN and Dropout sequentially (not shown in Fig.~\ref{ris:CNN_our}).
All the convolutional layers have equal kernel size 5 x 5. All the MaxPooling layers have equal kernel size 2 x 2, and stride 2. From now on, this CNN is marked as CNN-5.

\subsection{Performance metric}

For each class, binary classification problems are evaluated according to the principle of one versus all. Recall is used as a binary classification metric for each class. Recall is defined by the formula from \cite{olson2008advanced}:
\begin{equation}
Recall=\frac{TP}{TP+FN}
\end{equation}
where $TP$ is the number of samples when model correctly identified the considered class, $FN$ is the number of samples when model did not identify the considered class.

\subsection{Loss functions}
Binary Cross-Entropy is used as a loss function of the CNN-5:
\begin{equation}
BCE =-\frac{1}{N} \sum_{i=1}^{N} \cdot y_{i} \cdot \log \left(p\left(\hat{y_{i}}\right)\right)
+ \cdot \left(1-y_{i}\right) \cdot \log \left(1-p\left(\hat{y_{i}}\right)\right)
\end{equation}

\section{Results}
\label{sec:results}

Table~\ref{tab:comp1} presents the results of comparison for different preprocessing and feature engineering techniques and different CNN architectures for binary classification (normal pipe wall or defect/weld). Table~\ref{tab:comp2} shows the multiclass classification problem (normal pipe wall, defect or weld).

\begin{table}[!htb]
	\caption{\label{tab:comp1}Comparison of performance using Recall metric among different classification methods for binary classification problem. $y=0$ - healthy; $y=1$ - defect/weld}
	\begin{center}
		\small
		\begin{tabular}{lccc}
			\hline
			Method & $\hat{y}=y=0$ & $\hat{y}=y=1$ & Average \\
			\hline
			CNN-2 & 95.55 & 82.08 & 89.88 \\
			RayNet & 96.92 & 80.42 & 89.81  \\
			CNN-5 & 97.95 & \textbf{91.51} & \textbf{95.24} \\
			CNN-5+LRN & \textbf{98.29} & 89.86 & 94.74 \\
			\hline
			\multicolumn{4}{c}{Filling techniques comparison}  \\
			\hline
			CNN-5 (filling 1) & 97.95 & \textbf{91.51} & \textbf{95.24} \\
			CNN-5 (filling 2) & 97.95 & 84.20 & 92.16 \\
			CNN-5 (filling 3) & 97.26 & 83.02 & 91.27 \\
			CNN-5 (filling 4) & \textbf{98.63} & 81.13 & 91.27 \\
			CNN-5 (filling 5) & 98.12 & 81.84 & 91.27 \\
			\hline
		\end{tabular}
	\end{center}
\end{table}

\begin{table}[!h]
	\caption{\label{tab:comp2}Comparison of performance using Recall metric among different classification methods for multiclass classification problem. $y=0$ - healthy; $y=1$ - defect; $y=2$ - weld}
	\begin{center}
		\small
		\begin{tabular}{lcccc}
			\hline
			Method & $\hat{y}=y=0$ & $\hat{y}=y=1$ & $\hat{y}=y=2$ & Average \\
			\hline
			CNN-2 & 97.60 & 59.86 & 92.91 & 90.97 \\
			RayNet & \textbf{98.12} & \textbf{85.21} & 75.18 & 89.88  \\
			CNN-5 & \textbf{98.12} & 76.76 & \textbf{98.23} & \textbf{95.14} \\
			\hline
			\multicolumn{5}{c}{Single image normalization vs Whole dataset normalization}  \\
			\hline
			CNN-5 (1) (whole) & 97.95 & 64.08 & \textbf{99.65} & 93.65 \\
			CNN-5 (1) (image) &  98.12 & 76.76 & 98.23 & 95.14 \\
			CNN-2 (1) (whole) & 99.32 & 13.38 & 96.45 & 86.41 \\
			CNN-2 (1) (image) & 97.60 & 59.86 & 92.91 & 90.97 \\
			CNN-5 (3) (whole) & \textbf{99.66} & \textbf{81.69} & \textbf{99.65} & \textbf{97.12} \\
			CNN-2 (3) (whole) & 95.72 & 13.38 & 97.52 & 89.58 \\
			\hline
		\end{tabular}
	\end{center}
\end{table}

Batch size was equal to 64, so the input to the network had shape (64, 1, 64, 64). In the experiments we used the Adam optimizer with initial learning rate 0.001 and the learning rate scheduler with parameters: threshold = 0.0001, factor = 0.5, min lr = 0.0001, patience = 484. Also, for all the experiments, the number of epochs was equal to 12 and the dropout rate was equal to 0.33.

Filling methods were researched for binary classification problem. Centering means using a peak (extremums) searching procedure to define the weld or defect coordinates correctly. The centering procedure was researched for both the binary and multiclass classification problems. Moreover, the Min-Max normalization, with using either a single image or whole dataset, was investigated. Finally, CNN-2 and CNN-5 were compared for centered images with the first filling method using the single image Min-Max normalization.

\section{Conclusion}
\label{sec:conclusion}

Today, manual analysis of a magnetographic image is being a bottleneck for the diagnostics of pipelines, since it is costly limited by human resources. This study proves that this process can be fully automated, which is likely to make the analysis more reliable, faster and cheaper.

The CNN-5 network that outperformed the currently used CNNs for pipeline defect detection was proposed. Moreover, the results of the experiments prove that proper preprocessing procedures, including missing values filling techniques and normalization strategies, helps significantly improve the results and achieve the high quality of the oil pipeline diagnostics.

Finally, there can be several project development options:
\begin{enumerate}
	\item To increase sizes of the datasets,
	\item To improve the preprocessing procedures, including manual pictures selection,
	\item To try multiclass defects classification,
	\item To try defected and healthy welds classification,
	\item To apply defect depth evaluation,
	\item To investigate the repeatability of the results for similar datasets or transfer learning possibility.
\end{enumerate}

\bibliographystyle{unsrtnat}
\bibliography{mybibliography}

\end{document}